# Learning to Compose Task-Specific Tree Structures

**Jihun Choi, Kang Min Yoo, Sang-goo Lee**
Seoul National University, Seoul 08826, Korea
{jhchoi, kangminyoo, sglee}@europa.snu.ac.kr

## Abstract

For years, recursive neural networks (RvNNs) have been shown to be suitable for representing text into fixed-length vectors and achieved good performance on several natural language processing tasks. However, the main drawback of RvNNs is that they require structured input, which makes data preparation and model implementation hard. In this paper, we propose Gumbel Tree-LSTM, a novel tree-structured long short-term memory architecture that learns how to compose task-specific tree structures only from plain text data efficiently. Our model uses Straight-Through Gumbel-Softmax estimator to decide the parent node among candidates dynamically and to calculate gradients of the discrete decision. We evaluate the proposed model on natural language inference and sentiment analysis, and show that our model outperforms or is at least comparable to previous models. We also find that our model converges significantly faster than other models.

## Introduction

Techniques for mapping natural language into vector space have received a lot of attention, due to their capability of representing ambiguous semantics of natural language using dense vectors. Among them, methods of learning representations of words, e.g. word2vec (Mikolov et al. 2013) or GloVe (Pennington, Socher, and Manning 2014), are relatively well-studied empirically and theoretically (Baroni, Dinu, and Kruszewski 2014; Levy and Goldberg 2014), and some of them became typical choices to consider when initializing word representations for better performance at downstream tasks.

Meanwhile, research on sentence representation is still in active progress, and accordingly various architectures—designed with different intuition and tailored for different tasks—are being proposed. In the midst of them, three architectures are most frequently used in obtaining sentence representation from words. Convolutional neural networks (CNNs) (Kim 2014; Kalchbrenner, Grefenstette, and Blunsom 2014) utilize local distribution of words to encode sentences, similar to n-gram models. Recurrent neural networks (RNNs) (Dai and Le 2015; Kiros et al. 2015; Hill, Cho, and Korhonen 2016) encode sentences by reading words in sequential order. Recursive neural networks (RvNNs[1]) (Socher et al. 2013; Irsoy and Cardie 2014; Bowman et al. 2016), on which this paper focuses, rely on structured input (e.g. *parse tree*) to encode sentences, based on the intuition that there is significant semantics in the hierarchical structure of words. It is also notable that RvNNs are generalization of RNNs, as linear chain structures on which RNNs operate are equivalent to left- or right-skewed trees.

Although there is significant benefit in processing a sentence in a tree-structured recursive manner, data annotated with parse trees could be expensive to prepare and hard to be computed in batches (Bowman et al. 2016). Furthermore, the optimal hierarchical composition of words might differ depending on the properties of a task.

In this paper, we propose Gumbel Tree-LSTM, which is a novel RvNN architecture that does not require structured data and learns to compose task-specific tree structures without explicit guidance. Our Gumbel Tree-LSTM model is based on tree-structured long short-term memory (Tree-LSTM) architecture (Tai, Socher, and Manning 2015; Zhu, Sobihani, and Guo 2015), which is one of the most renowned variants of RvNN.

To learn how to compose task-specific tree structures without depending on structured input, our model introduces *composition query vector* that measures validity of a composition. Using validity scores computed by the composition query vector, our model recursively selects compositions until only a single representation remains. We use Straight-Through (ST) Gumbel-Softmax estimator (Jang, Gu, and Poole 2017; Maddison, Mnih, and Teh 2017) to sample compositions in the training phase. ST Gumbel-Softmax estimator relaxes the discrete sampling operation to be continuous in the backward pass, thus our model can be trained via the standard backpropagation. Also, since the computation is performed layer-wise, our model is easy to implement and naturally supports batched computation.

From experiments on natural language inference and sentiment analysis tasks, we find that our proposed model outperforms or is at least comparable to previous sentence encoder models and converges significantly faster than them.

The contributions of our work are as follows:

---

[1]In some RvNN papers, the term 'recursive neural network' is often abbreviated to 'RNN', however to avoid confusion with recurrent neural network we decided to use the acronym 'RvNN'.

- We designed a novel sentence encoder architecture that learns to compose task-specific trees from plain text data.
- We showed from experiments that the proposed architecture outperforms or is competitive to state-of-the-art models. We also observed that our model converges faster than others.
- Specifically, we saw that our model significantly outperforms previous RvNN works trained on parse trees in all conducted experiments, from which we hypothesize that syntactic parse tree may not be the best structure for every task and the optimal structure could differ per task.

In the next section, we briefly introduce previous works which have similar objectives to that of our work. Then we describe the proposed model in detail and present findings from experiments. Lastly we summarize the overall content and discuss future work.

## Related Work

There have been several works that aim to learn hierarchical latent structure of text by recursively composing words into sentence representation. Some of them carry unsupervised learning on structures by making composition operations soft. To the best of our knowledge, gated recursive convolutional neural network (grConv) (Cho et al. 2014) is the first model of its kind and used as an encoder for neural machine translation. The grConv architecture uses gating mechanism to control the information flow from children to parent. grConv and its variants are also applied to sentence classification tasks (Chen et al. 2015; Zhao, Lu, and Poupart 2015). Neural tree indexer (NTI) (Munkhdalai and Yu 2017b) utilizes soft hierarchical structures by using Tree-LSTM instead of grConv.

Although models that operate with soft structures are naturally capable of being trained via backpropagation, the structures predicted by them are ambiguous and thus it is hard to interpret them. CYK Tree-LSTM (Maillard, Clark, and Yogatama 2017) resolves this ambiguity while maintaining the soft property by introducing the concept of CYK parsing algorithm (Kasami 1965; Younger 1967; Cocke 1970). Though their model reduces the ambiguity by explicitly representing a node as a weighted sum of all candidate compositions, it is memory intensive since the number of candidates linearly increases by depth.

On the other hand, there exist some previous works that maintain the discreteness of tree composition processes, instead of relying on the soft hierarchical structure. The architecture proposed by Socher et al. (2011) greedily selects two adjacent nodes whose reconstruction error is the smallest and merges them into the parent. In their work, rather than directly optimized on classification loss, a composition function is optimized to minimize reconstruction error.

Yogatama et al. (2017) introduce reinforcement learning to achieve the desired effect of discretization. They show that REINFORCE (Williams 1992) algorithm can be used in estimating gradients to learn a tree composition function minimizing classification error. However, slow convergence due to the reinforcement learning setting is one of its drawbacks, according to the authors.

In the research area outside the RvNN, compositionality in vector space also has been a longstanding subject (Plate 1995; Mitchell and Lapata 2010; Grefenstette and Sadrzadeh 2011; Zanzotto and Dell'Arciprete 2012, to name a few). And more recently, there exist works aiming to learn hierarchical latent structure from unstructured data (Chung, Ahn, and Bengio 2017; Kim et al. 2017).

## Model Description

Our proposed architecture is built based on the tree-structured long short-term memory network architecture. We introduce several additional components into the Tree-LSTM architecture to allow the model to dynamically compose tree structure in a bottom-up manner and to effectively encode a sentence into a vector. In this section, we describe the components of our model in detail.

### Tree-LSTM

Tree-structured long short-term memory network (Tree-LSTM) (Tai, Socher, and Manning 2015; Zhu, Sobihani, and Guo 2015) is an elegant variant of RvNN, where it controls information flow from children to parent using similar mechanism to long short-term memory (LSTM) (Hochreiter and Schmidhuber 1997). Tree-LSTM introduces *cell state* in computing parent representation, which assists each cell to capture distant vertical dependencies.

The following are formulae that our model uses to compute parent representation from its children:

$$\begin{bmatrix} \mathbf{i} \\ \mathbf{f}_l \\ \mathbf{f}_r \\ \mathbf{o} \\ \mathbf{g} \end{bmatrix} = \begin{bmatrix} \sigma \\ \sigma \\ \sigma \\ \sigma \\ \tanh \end{bmatrix} \left( \mathbf{W}_{comp} \begin{bmatrix} \mathbf{h}_l \\ \mathbf{h}_r \end{bmatrix} + \mathbf{b}_{comp} \right) \quad (1)$$

$$\mathbf{c}_p = \mathbf{f}_l \odot \mathbf{c}_l + \mathbf{f}_r \odot \mathbf{c}_r + \mathbf{i} \odot \mathbf{g} \quad (2)$$

$$\mathbf{h}_p = \mathbf{o} \odot \tanh(\mathbf{c_p}), \quad (3)$$

where $\mathbf{W}_{comp} \in \mathbb{R}^{5D_h \times 2D_h}$ $\mathbf{b}_{comp} \in \mathbb{R}^{2D_h}$, and $\odot$ is the element-wise product. Note that our formulation is akin to that of SPINN (Bowman et al. 2016), but our version does not include the tracking LSTM. Instead, our model can apply an LSTM to leaf nodes, which we will soon describe.

### Gumbel-Softmax

Gumbel-Softmax (Jang, Gu, and Poole 2017) (or Concrete distribution (Maddison, Mnih, and Teh 2017)) is a method of utilizing discrete random variables in a network. Since it approximates one-hot vectors sampled from a categorical distribution by making them continuous, gradients of model parameters can be calculated using the reparameterization trick and the standard backpropagation. Gumbel-Softmax is known to have an advantage over score-function-based gradient estimators such as REINFORCE (Williams 1992) which suffer from high variance and slow convergence (Jang, Gu, and Poole 2017).

Gumbel-Softmax distribution is motivated by Gumbel-Max trick (Maddison, Tarlow, and Minka 2014), an algorithm for sampling from a categorical distribution. Consider a $k$-dimensional categorical distribution whose class probabilities $p_1, \cdots, p_k$ are defined in terms of unnormalized log probabilities $\pi_1, \cdots, \pi_k$:

$$p_i = \frac{\exp(\log(\pi_i))}{\sum_{j=1}^{k} \exp(\log(\pi_j))}. \tag{4}$$

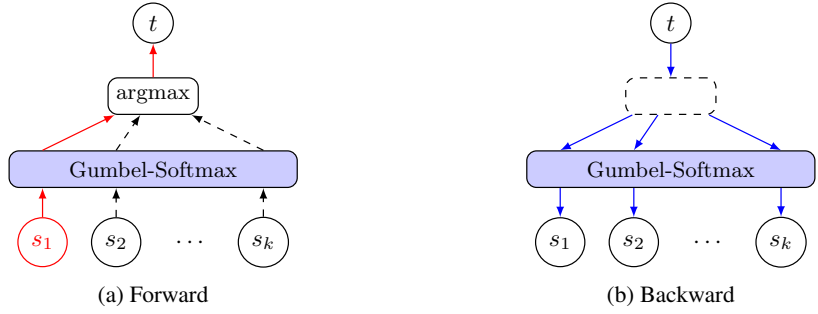

Figure 1: Visualization of forward and backward computation path of ST Gumbel-Softmax. In the forward pass, a model can maintain sparseness due to $\arg\max$ operation. In the backward pass, since there is no discrete operation, the error signal can backpropagate.

Then a one-hot sample $\mathbf{z} = (z_1, \cdots, z_k) \in \mathbb{R}^k$ from the distribution can be easily drawn by the following equations:

$$z_i = \begin{cases} 1 & i = \arg\max_j(\log(\pi_j) + g_j) \\ 0 & \text{otherwise} \end{cases} \tag{5}$$

$$g_i = -\log(-\log(u_i)) \tag{6}$$

$$u_i \sim \text{Uniform}(0,1). \tag{7}$$

Here, $g_i$, namely *Gumbel noise*, perturbs each $\log(\pi_i)$ term so that taking $\arg\max$ becomes equivalent to drawing a sample weighted on $p_1, \cdots, p_k$.

In Gumbel-Softmax, the discontinuous $\arg\max$ function of Gumbel-Max trick is replaced by the differentiable softmax function. That is, given unnormalized probabilities $\pi_1, \cdots, \pi_k$, a sample $\mathbf{y} = (y_1, \cdots, y_k)$ from the Gumbel-Softmax distribution is drawn by

$$y_i = \frac{\exp((\log(\pi_i) + g_i)/\tau)}{\sum_{j=1}^{k} \exp((\log(\pi_j) + g_j)/\tau)}, \tag{8}$$

where $\tau$ is a temperature parameter; as $\tau$ diminishes to zero, a sample from the Gumbel-Softmax distribution becomes *cold* and resembles the one-hot sample.

Straight-Through (ST) Gumbel-Softmax estimator (Jang, Gu, and Poole 2017), whose name reminds of Straight-Through estimator (STE) (Bengio, Léonard, and Courville 2013), is a discrete version of the continuous Gumbel-Softmax estimator. Similar to the STE, it maintains sparsity by taking different paths in the forward and backward propagation. Obviously ST estimators are biased, however they perform well in practice, according to several previous works (Chung, Ahn, and Bengio 2017; Gu, Im, and Li 2017) and our own result.

In the forward pass, it discretizes a continuous probability vector $\mathbf{y}$ sampled from the Gumbel-Softmax distribution into the one-hot vector $\mathbf{y}^{ST} = (y_1^{ST}, \cdots, y_k^{ST})$, where

$$y_i^{ST} = \begin{cases} 1 & i = \arg\max_j y_j \\ 0 & \text{otherwise} \end{cases}. \tag{9}$$

And in the backward pass it simply uses the continuous $\mathbf{y}$, thus the error signal is still able to backpropagate. See Figure 1 for the visualization of the forward and backward pass.

ST Gumbel-Softmax estimator is useful when a model needs to utilize discrete values directly, for example in the case that a model alters its computation path based on samples drawn from a categorical distribution.

**Gumbel Tree-LSTM**

In our Gumbel Tree-LSTM model, an input sentence composed of $N$ words is represented as a sequence of word vectors $(\mathbf{x}_1, \cdots, \mathbf{x}_N)$, where $\mathbf{x}_i \in \mathbb{R}^{D_x}$. Our basic model applies an affine transformation to each $\mathbf{x}_i$ to obtain the initial hidden and cell state:

$$\mathbf{r}_i^1 = \begin{bmatrix} \mathbf{h}_i^1 \\ \mathbf{c}_i^1 \end{bmatrix} = \mathbf{W}_{leaf} \mathbf{x}_i + \mathbf{b}_{leaf}, \tag{10}$$

which we call *leaf transformation*. In Eq. 10, $\mathbf{W}_{leaf} \in \mathbb{R}^{2D_h \times D_x}$ and $\mathbf{b}_{leaf} \in \mathbb{R}^{2D_h}$. Note that we denote the representation of $i$-th node at $t$-th layer as $\mathbf{r}_i^t = [\mathbf{h}_i^t; \mathbf{c}_i^t]$.

Assume that $t$-th layer consists of $M_t$ node representations: $(\mathbf{r}_1^t, \cdots, \mathbf{r}_{M_t}^t)$. If two adjacent nodes, say $\mathbf{r}_i^t$ and $\mathbf{r}_{i+1}^t$, are selected to be merged, then Eqs. 1–3 are applied by assuming $[\mathbf{h}_l; \mathbf{c}_l] = \mathbf{r}_i^t$ and $[\mathbf{h}_r; \mathbf{c}_r] = \mathbf{r}_{i+1}^t$ to obtain the parent representation $[\mathbf{h}_p; \mathbf{c}_p] = \mathbf{r}_i^{t+1}$. Node representations which are not selected are copied to the corresponding positions at layer $t+1$. In other words, the $(t+1)$-th layer is composed of $M_{t+1} = M_t - 1$ representations $(\mathbf{r}_1^{t+1}, \cdots, \mathbf{r}_{M_{t+1}}^{t+1})$, where

$$\mathbf{r}_j^{t+1} = \begin{cases} \mathbf{r}_j^t & j < i \\ \text{Tree-LSTM}(\mathbf{r}_j^t, \mathbf{r}_{j+1}^t) & j = i \\ \mathbf{r}_{j+1}^t & j > i \end{cases}. \tag{11}$$

This procedure is repeated until the model reaches $N$-th layer and only a single node is left. It is notable that the property of selecting the best node pair at each stage resembles that of easy-first parsing (Goldberg and Elhadad 2010). For implementation-wise details, please see the supplementary material.

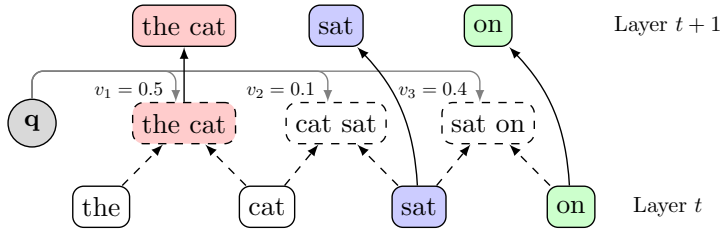

Figure 2: An example of the parent selection. At layer $t$ (the bottom layer), the model computes parent candidates (the middle layer). Then the validity score of each candidate is computed using the query vector $\mathbf{q}$ (denoted as $v_1, v_2, v_3$). In the training time, the model samples a parent node among candidates weighted on $v_1, v_2, v_3$, using ST Gumbel-Softmax estimator, and in the testing time the model selects the candidate with the highest validity. At layer $t+1$ (the top layer), the representation of the selected candidate ('the cat') is used as a parent, and the rest are copied from those of layer $t$ ('sat', 'on'). Best viewed in color.

**Parent selection.** Since information about the tree structure of an input is not given to the model, a special mechanism is needed for the model to learn to compose task-specific tree structures in an end-to-end manner. We now describe the mechanism for building up the tree structure from an unstructured sentence.

First, our model introduces the trainable *composition query vector* $\mathbf{q} \in \mathbb{R}^{D_h}$. The composition query vector measures how valid a representation is. Specifically, the *validity score* of a representation $\mathbf{r} = [\mathbf{h}; \mathbf{c}]$ is defined by $\mathbf{q} \cdot \mathbf{h}$.

At layer $t$, the model computes candidates for the parent representations using Eqs. 1–3: $(\tilde{\mathbf{r}}_1^{t+1}, \cdots, \tilde{\mathbf{r}}_{M_{t+1}}^{t+1})$. Then, it calculates the validity score of each candidate and normalize it so that $\sum_{i=1}^{M_{t+1}} v_i = 1$:

$$v_i = \frac{\exp(\mathbf{q} \cdot \tilde{\mathbf{h}}_i^{t+1})}{\sum_{j=1}^{M_{t+1}} \exp(\mathbf{q} \cdot \tilde{\mathbf{h}}_j^{t+1})}. \quad (12)$$

In the training phase, the model samples a parent from candidates weighted on $v_i$, using the ST Gumbel-Softmax estimator described above. Since the continuous Gumbel-Softmax function is used in the backward pass, the error backpropagation signal safely passes through the sampling operation, hence the model is able to learn to construct the task-specific tree structures that minimize the loss by back-propagation.

In the validation (or testing) phase, the model simply selects the parent which maximizes the validity score.

An example of the parent selection is depicted in Figure 2.

**LSTM-based leaf transformation.** The basic leaf transformation using an affine transformation (Eq. 10) does not consider information about the entire sentence of an input and thus the parent selection is performed based only on local information.

SPINN (Bowman et al. 2016) addresses this issue by using the tracking LSTM which sequentially reads input words. The tracking LSTM makes the SPINN model *hybrid*, where the model takes advantage of both tree-structured composition and sequential reading. However, the tracking LSTM is not applicable to our model, since our model does not use shift-reduce parsing or maintain a stack.

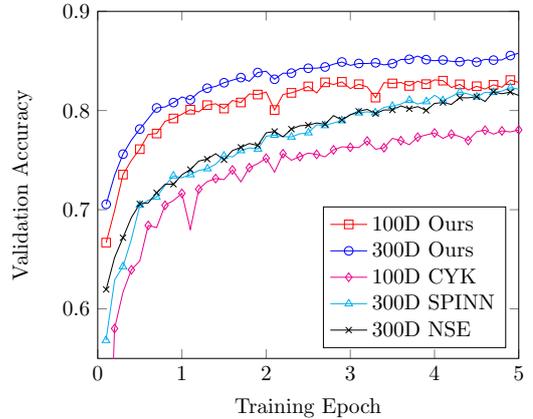

Figure 3: Validation accuracies during training.

In the tracking LSTM's stead, our model applies an LSTM on input representations to give information about previous words to each leaf node:

$$\mathbf{r}_i^1 = \begin{bmatrix} \mathbf{h}_i^1 \\ \mathbf{c}_i^1 \end{bmatrix} = \text{LSTM}(\mathbf{x_i}, \mathbf{h}_{i-1}^1, \mathbf{c}_{i-1}^1), \quad (13)$$

where $\mathbf{h}_0^1 = \mathbf{c}_0^1 = \vec{0}$.

From the experimental results, we validate that the LSTM applied to leaf nodes has a substantial gain over the basic leaf transformer.

## Experiments

We evaluate performance of the proposed Gumbel Tree-LSTM model on two tasks: natural language inference and sentiment analysis. The implementation is made publicly available.[2] The detailed experimental settings are described in the supplementary material.

### Natural Language Inference

Natural language inference (NLI) is a task of predicting the relationship between two sentences (hypothesis and premise). In the Stanford Natural Language Inference (SNLI) dataset (Bowman et al. 2015), which we use for NLI experiments, a relationship is either contradiction, entailment, or neutral. For a model to correctly predict the relationship between two sentences, it should encode semantics of sentences accurately, thus the task has been used as one of standard tasks for evaluating the quality of sentence representations.

---
[2] https://github.com/jihunchoi/unsupervised-treelstm

| Model | Accuracy (%) | # Params | Time (hours) |
|---|---|---|---|
| 100D Latent Syntax Tree-LSTM (Yogatama et al. 2017) | 80.5 | 500k | 72–96* |
| 100D CYK Tree-LSTM (Maillard, Clark, and Yogatama 2017) | 81.6 | 231k | 240* |
| 100D Gumbel Tree-LSTM, without Leaf LSTM (Ours) | 81.8 | 202k | 0.7 |
| 100D Gumbel Tree-LSTM (Ours) | **82.6** | 262k | **0.6** |
| 300D LSTM (Bowman et al. 2016) | 80.6 | 3.0M | 4† |
| 300D SPINN (Bowman et al. 2016) | 83.2 | 3.7M | 67† |
| 300D NSE (Munkhdalai and Yu 2017a) | 84.6 | 3.0M | 26† |
| 300D Gumbel Tree-LSTM, without Leaf LSTM (Ours) | 84.4 | 2.3M | 3.1 |
| 300D Gumbel Tree-LSTM (Ours) | **85.6** | 2.9M | **1.6** |
| 600D (300+300) Gated-Attention BiLSTM (Chen et al. 2017) | 85.5 | 11.6M | 8.5† |
| 512–1024–2048D Shortcut-Stacked BiLSTM (Nie and Bansal 2017) | **86.1** | 140.2M | 3.8†‡ |
| 600D Gumbel Tree-LSTM (Ours) | 86.0 | 10.3M | **3.4** |

Table 1: Results of SNLI experiments. The above two sections group models of similar numbers of parameters. The bottom section contains results of state-of-the-art models. Word embedding parameters are not included in the number of parameters. ∗: values reported in the original papers. †: values estimated from per-epoch training time on the same machine our models trained on. ‡: cuDNN library is used in RNN computation.

The SNLI dataset is composed of about 550,000 sentences, each of which is binary-parsed. However, since our model operate on plain text, we do not use the parse tree information in both training and testing. The classifier architecture used in our SNLI experiments follows (Mou et al. 2016; Chen et al. 2017). Given the premise sentence vector ($\mathbf{h}^{pre}$) and the hypothesis sentence vector ($\mathbf{h}^{hyp}$) which are encoded by the proposed Gumbel Tree-LSTM model, the probability of relationship $r \in$ {entailment, contradiction, neutral} is computed by the following equations:

$$p(r|\mathbf{h}^{pre}, \mathbf{h}^{hyp}) = \text{softmax}(\mathbf{W}^r_{clf}\mathbf{a} + \mathbf{b}^r_{clf}) \quad (14)$$

$$\mathbf{a} = \Phi(\mathbf{f}) \quad (15)$$

$$\mathbf{f} = \begin{bmatrix} \mathbf{h}^{pre} \\ \mathbf{h}^{hyp} \\ |\mathbf{h}^{pre} - \mathbf{h}^{hyp}| \\ \mathbf{h}^{pre} \odot \mathbf{h}^{hyp} \end{bmatrix}, \quad (16)$$

where $\mathbf{W}^r_{clf} \in \mathbb{R}^{1 \times D_c}$, $\mathbf{b}^r_{clf} \in \mathbb{R}^1$, and $\Phi$ is a multi-layer perceptron (MLP) with the rectified linear unit (ReLU) activation function.

For 100D experiments (where $D_x = D_h = 100$), we use a single-hidden layer MLP with 200 hidden units (i.e. $D_c = 200$. The word vectors are initialized with GloVe (Pennington, Socher, and Manning 2014) 100D pretrained vectors[3] and fine-tuned during training.

For 300D experiments (where $D_x = D_h = 300$), we set the number of hidden units of a single-hidden layer MLP to 1024 ($D_c = 1024$) and added batch normalization layers (Ioffe and Szegedy 2015) followed by dropout (Srivastava et al. 2014) with probability 0.1 to the input and the output of the MLP. We also apply dropout on the word vectors with probability 0.1. Similar to 100D experiments, we initialize the word embedding matrix with GloVe 300D pretrained vectors[4], however we do not update the word representations during training.

Since our model converges relatively fast, it is possible to train a model of larger size in a reasonable time. In the 600D experiment, we set $D_x = 300$, $D_h = 600$, and an MLP with three hidden layers ($D_c = 1024$) is used. The dropout probability is set to 0.2 and word embeddings are not updated during training.

The size of mini-batches is set to 128 in all experiments, and hyperparameters are tuned using the validation split. The temperature parameter $\tau$ of Gumbel-Softmax is set to 1.0, and we did not find that temperature annealing improves performance. For training models, Adam optimizer (Kingma and Ba 2015) is used.

The results of SNLI experiments are summarized in Table 1. First, we can see that LSTM-based leaf transformation has a clear advantage over the affine-transformation-based one. It improves the performance substantially and also leads to faster convergence.

Secondly, comparing ours with other models, we find that our 100D and 300D model outperform all other models of similar numbers of parameters. Our 600D model achieves the accuracy of 86.0%, which is comparable to that of the state-of-the-art model (Nie and Bansal 2017), while using far less parameters.

It is also worth noting that our models converge much faster than other models. All of our models converged within a few hours on a machine with NVIDIA Titan Xp GPU.

We also plot validation accuracies of various models during first 5 training epochs in Figure 3, and validate that our models converge significantly faster than others, not only in terms of total training time but also in the number of iterations.[5]

---
[3] http://nlp.stanford.edu/data/glove.6B.zip

[4] http://nlp.stanford.edu/data/glove.840B.300d.zip

| Model | SST-2 (%) | SST-5 (%) |
|---|---|---|
| DMN (Kumar et al. 2016) | 88.6 | 52.1 |
| NSE (Munkhdalai and Yu 2017a) | 89.7 | 52.8 |
| byte-mLSTM (Radford, Jozefowicz, and Sutskever 2017) | **91.8** | 52.9 |
| BCN+Char+CoVe (McCann et al. 2017) | 90.3 | **53.7** |
| RNTN (Socher et al. 2013) | 85.4 | 45.7 |
| Constituency Tree-LSTM (Tai, Socher, and Manning 2015) | 88.0 | 51.0 |
| NTI-SLSTM-LSTM (Munkhdalai and Yu 2017b) | 89.3 | 53.1 |
| Latent Syntax Tree-LSTM (Yogatama et al. 2017) | 86.5 | – |
| Constituency Tree-LSTM + Recurrent Dropout (Looks et al. 2017) | 89.4 | 52.3 |
| Gumbel Tree-LSTM (Ours) | <u>90.7</u> | <u>53.7</u> |

Table 2: Results of SST experiments. The bottom section contains results of RvNN-based models. Underlined score indicates the best among RvNN-based models.

### Sentiment Analysis

To evaluate the performance of our model in single-sentence classification, we conducted experiments on Stanford Sentiment Treebank (SST) (Socher et al. 2013) dataset. In the SST dataset, each sentence is represented as a binary parse tree, and each subtree of a parse tree is annotated with the corresponding sentiment score. Following the experimental setting of previous works, we use all subtrees and their labels for training, and only the root labels are used for evaluation.

The classifier has a similar architecture to SNLI experiments. Specifically, for a sentence embedding $\mathbf{h}$, the probability for the sentence to be predicted as label $s \in \{0, 1\}$ (in the binary setting, SST-2) or $s \in \{1, 2, 3, 4, 5\}$ (in the fine-grained setting, SST-5) is computed as follows:

$$p(s|\mathbf{h}) = \text{softmax}(\mathbf{W}_{clf}^s \mathbf{a} + \mathbf{b}_{clf}^s) \quad (17)$$

$$\mathbf{a} = \Phi(\mathbf{h}), \quad (18)$$

where $\mathbf{W}_{clf}^s \in \mathbb{R}^{1 \times D_c}$, $\mathbf{b}_{clf}^s \in \mathbb{R}^1$, and $\Phi$ is a single-hidden layer MLP with the ReLU activation function. Note that subtrees labeled as neutral are ignored in the binary setting in both training and evaluation.

We trained our SST-2 model with hyperparameters $D_x = 300$, $D_h = 300$, $D_c = 300$. The word vectors are initialized with GloVe 300D pretrained vectors and fine-tuned during training. We apply dropout ($p = 0.5$) on the output of the word embedding layer and the input and the output of the MLP layer. The size of mini-batches is set to 32 and Adadelta (Zeiler 2012) optimizer is used for optimization.

For our SST-5 model, hyperparameters are set to $D_x = 300$, $D_h = 300$, $D_c = 1024$. Similar to the SST-2 model, we optimize the model using Adadelta optimizer with batch size 64 and apply dropout with $p = 0.5$.

Table 2 summarizes the results of SST experiments. Our SST-2 model outperforms all other models substantially except byte-mLSTM (Radford, Jozefowicz, and Sutskever 2017), where a byte-level language model trained on the large product review dataset is used to obtain sentence representations.

---
[5] In the figure, our models and 300D NSE are trained with batch size 128. 100D CYK and 300D SPINN are trained with batch size 16 and 32 respectively, as in the original papers. We observed that our models still converge faster than others when a smaller batch size (16 or 32) is used.

We also see that the performance of our SST-5 model is on par with that of the current state-of-the-art model (McCann et al. 2017), which is pretrained on large parallel datasets and uses character n-gram embeddings alongside word embeddings, even though our model does not utilize external resources other than GloVe vectors and only uses word-level representations. The authors of (McCann et al. 2017) stated that utilizing pretraining and character n-gram embeddings improves validation accuracy by 2.8% (SST-2) or 1.7% (SST-5).

In addition, from the fact that our models substantially outperform all other RvNN-based models, we conjecture that task-specific tree structures built by our model help encode sentences into vectors more efficiently than constituency-based or dependency-based parse trees do.

### Qualitative Analysis

We conduct a set of experiments to observe various properties of our trained models. First, to see how well the model encodes sentences with similar meaning or syntax into close vectors, we find nearest neighbors of a query sentence. Second, to validate that the trained composition functions are non-trivial and task-specific, we visualize trees composed by SNLI and SST model given identical sentence.

**Nearest neighbors** We encode sentences in the test split of SNLI dataset using the trained 300D model and find nearest neighbors given a query sentence. Table 3 presents five nearest neighbors for each selected query sentence. In finding nearest neighbors, cosine distance is used as metric. The result shows that our model effectively maps similar sentences into vectors close to each other; the neighboring sentences are similar to a query sentence not only in terms of word overlap, but also in semantics. For example in the second column, the nearest sentence is 'the woman is looking at a dog', whose meaning is almost same as the query sentence. We can also see that other neighbors partially share semantics with the query sentence.

| # | sunshine is on a man 's face . | a girl is staring at a dog . | the woman is wearing boots . |
|---|---|---|---|
| 1 | a man is walking on sunshine . | the woman is looking at a dog . | the girl is wearing shoes |
| 2 | a guy is in a hot , sunny place | a girl takes a photo of a dog . | a person is wearing boots . |
| 3 | a man is working in the sun . | a girl is petting her dog . | the woman is wearing jeans . |
| 4 | it is sunny . | a man is taking a picture of a dog , while a woman watches . | a woman wearing sunglasses . |
| 5 | a man enjoys the sun coming through the window . | a woman is playing with her dog . | the woman is wearing a vest . |

Table 3: Nearest neighbor sentences of query sentences. Each query sentence is unseen in the dataset.

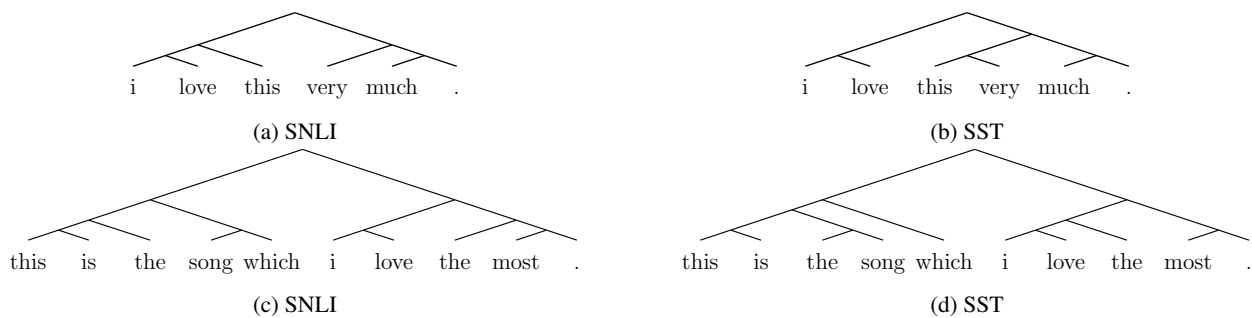

Figure 4: Tree structures built by models trained on SNLI and SST.

**Tree examples** Figure 4 show that two models (300D SNLI and SST-2) generate different tree structures given an identical sentence. In Figure 4a and 4b, the SNLI model groups the phrase 'i love this' first, while the SST model groups 'this very much' first. Figure 4c and 4d present how differently the two models process a sentence containing relative pronoun 'which'. It is intriguing that the models compose visually plausible tree structures, where the sentence is divided into two phrases by relative pronoun, even though they are trained without explicit parse trees. We hypothesize that these examples demonstrate that each model generates a distinct tree structure based on semantic properties of the task and learns non-trivial tree composition scheme.

## Conclusion

In this paper, we propose Gumbel Tree-LSTM, a novel Tree-LSTM-based architecture that learns to compose task-specific tree structures. Our model introduces the composition query vector to compute validity of the candidate parents and selects the appropriate parent according to validity scores. In training time, the model samples the parent from candidates using ST Gumbel-Softmax estimator, hence it is able to be trained by standard backpropagation while maintaining its property of discretely determining the computation path in forward propagation.

From experiments, we validate that our model outperforms all other RvNN models and is competitive to state-of-the-art models, and also observed that our model converges faster than other complex models. The result poses an important question: *what is the optimal input structure for RvNN?* We empirically showed that the optimal structure might differ per task, and investigating task-specific latent tree structures could be an interesting future research direction.

For future work, we plan to apply the core idea beyond sentence encoding. The performance could be further improved by applying intra-sentence or inter-sentence attention mechanisms. We also plan to design an architecture that generates sentences using recursive structures.

## Appendix

The supplementary material is available at https://github.com/jihunchoi/unsupervised-treelstm/blob/master/aaai18/supp.pdf.

## Acknowledgments

This work is part of SNU-Samsung smart campus research program, which is supported by Samsung Electronics. The authors would like to thank anonymous reviewers for valuable comments and Volkan Cirik for helpful feedback on the early version of the manuscript.

# Supplementary Material for "Learning to Compose Task-Specific Tree Structures"

**Jihun Choi, Kang Min Yoo, Sang-goo Lee**
Seoul National University, Seoul 08826, Korea
{jhchoi, kangminyoo, sglee}@europa.snu.ac.kr

## Implementation Details

Implementation-wise, we used multiple *mask* matrices in implementing the proposed Gumbel Tree-LSTM model. Using the mask matrices, Eq. 11 can be rewritten as a single equation:

$$\mathbf{r}^{t+1}_{1:M_{t+1}} = \mathbf{M}_l \odot \mathbf{r}^t_{1:M_t-1} + \mathbf{M}_r \odot \mathbf{r}^t_{2:M_t} + \mathbf{M}_p \odot \tilde{\mathbf{r}}^{t+1}_{1:M_{t+1}}. \quad (S1)$$

In the above equation, $\mathbf{M}_l, \mathbf{M}_r, \mathbf{M}_p \in \mathbb{R}^{D_h \times M_{t+1}}$, and $\mathbf{r}^t_{1:L} \in \mathbb{R}^{D_h \times L}$ is a matrix whose columns are $\mathbf{r}^t_1, \cdots, \mathbf{r}^t_L \in \mathbb{R}^{D_h}$.

The mask matrices are defined by the following equations.

$$\mathbf{M}_l = [\mathbf{m}_l \ \cdots \ \mathbf{m}_l]^T \quad (S2)$$

$$\mathbf{M}_r = [\mathbf{m}_r \ \cdots \ \mathbf{m}_r]^T \quad (S3)$$

$$\mathbf{M}_p = [\mathbf{m}_p \ \cdots \ \mathbf{m}_p]^T \quad (S4)$$

$$\mathbf{m}_l = \mathbf{1} - \texttt{cumsum}(\bar{\mathbf{y}}_{1:M_{t+1}}) \quad (S5)$$

$$\mathbf{m}_r = [0 \ \ \texttt{cumsum}(\bar{\mathbf{y}}_{1:M_{t+1}-1})]^T \quad (S6)$$

$$\mathbf{m}_p = \bar{\mathbf{y}}_{1:M_{t+1}} \quad (S7)$$

Here, $\texttt{cumsum}(\mathbf{c})$ is a function that takes a vector $\mathbf{c} = [c_1 \cdots c_k]^T$ and outputs a vector $\mathbf{d} = [d_1 \cdots d_k]^T$ s.t. $d_i = \sum_{j=1}^{i} c_j$. $\bar{\mathbf{y}}_{1:M_{t+1}} \in \mathbb{R}^{M_{t+1}}$ is a vector which will be defined below, and $\mathbf{1} \in \mathbb{R}^{M_{t+1}}$ is a vector whose values are all ones.

In the forward pass, $\bar{\mathbf{y}}_{1:M_{t+1}}$ is defined by a one-hot vector $\mathbf{y}^{ST}_{1:M_{t+1}}$, which is sampled from the categorical distribution of validity scores $(v_1, \cdots, v_{M_{t+1}})$ using Gumbel-Max trick.

$$y^{ST}_i = \begin{cases} 1 & i = \arg\max_j \left( \mathbf{q} \cdot \tilde{\mathbf{h}}^{t+1}_j + g_j \right) \\ 0 & \text{otherwise} \end{cases} \quad (S8)$$

$$g_i = -\log(-\log(u_i + \epsilon) + \epsilon) \quad (S9)$$

$$u_i \sim \text{Uniform}(0,1) \quad (S10)$$

Note that $\epsilon = 10^{-20}$ is added when calculating $g_i$ for numerical stability.

In the backward pass, instead of the one-hot version, the continuous vector $\mathbf{y}_{1:M_{t+1}}$ obtained from Gumbel-Softmax is used as $\bar{\mathbf{y}}_{1:M_{t+1}}$. Note that the Gumbel noise samples $g_1, \cdots, g_{M_{t+1}}$ drawn in the forward pass are reused in the backward pass (i.e. noise values are not resampled in the backward pass).

In typical deep learning libraries supporting automatic differentiation (e.g. PyTorch, TensorFlow), this discrepancy between forward and backward pass can be implemented as

$$\bar{\mathbf{y}}_{1:M_{t+1}} = \texttt{detach}(\mathbf{y}^{ST}_{1:M_{t+1}} - \mathbf{y}_{1:M_{t+1}}) + \mathbf{y}_{1:M_{t+1}}, \quad (S11)$$

where $\texttt{detach}(\cdot)$ is a function that prevents error from backpropagating through its input.

## Detailed Experimental Settings

All experiments are conducted using the publicized codebase.[1]

### SNLI

The composition query vector is initialized by sampling from Gaussian distribution $\mathcal{N}(0, 0.01^2)$. The last linear transformation that outputs the unnormalized log probability for each class is initialized by sampling from uniform distribution $\mathcal{U}(-0.005, 0.005)$. All other parameters are initialized following the scheme proposed by He et al. (2015). We used Adam optimizer (Kingma and Ba 2015) with default hyperparameters and halved learning rate if there is no improvement in accuracy for one epoch. The size of mini-batch is set to 128 in all experiments.

In 100D experiments ($D_x = D_h = 100$, $D_c = 200$, single-hidden layer MLP classifier), GloVe (6B, 100D) (Pennington, Socher, and Manning 2014) pretrained word embeddings are used in initializing word representations. We fine-tuned word embedding parameters during training.

In 300D ($D_x = D_h = 300$, $D_c = 1024$, single-hidden layer MLP classifier) and 600D ($D_x = 300$, $D_h = 600$, $D_c = 1024$, MLP classifier with three hidden layers) experiments, GloVe (840B, 300D) pretrained word embeddings are used as word representations and fixed during training. Batch normalization is applied before the input and after the output of the MLP. Dropout is applied to word embeddings and the input and the output of the MLP with dropout probability 0.1 (300D) or 0.2 (600D).

---

[1] https://github.com/jihunchoi/unsupervised-treelstm

**SST**

The composition query vector is initialized by sampling from Gaussian distribution $\mathcal{N}(0, 0.01^2)$. The last linear transformation that outputs the unnormalized log probability for each class is initialized by sampling from uniform distribution $\mathcal{U}(-0.002, 0.002)$. All other parameters are initialized following the scheme proposed by He et al. (2015). We used Adadelta optimizer (Zeiler 2012) with default hyperparameters and halved learning rate if there is no improvement in accuracy for two epochs. In both SST-2 and SST-5 experiments, we set $D_x = D_h = 300$, used GloVe (840B, 300D) pretrained vectors with fine-tuning, and single-hidden layer MLP is used as classifier. Dropout is applied to word embeddings and the input and the output of the MLP classifier with probability 0.5.

In the SST-2 experiment, we set $D_c$ to 300 and set batch size to 32. In the SST-5 experiment, $D_c$ is increased to 1024, and mini-batches of 64 sentences are fed to the model during training.